\documentclass[10pt,a4paper]{article}

\usepackage[a4paper,left=25mm,right=19mm,top=20mm,bottom=22mm]{geometry}
\usepackage{mathptmx}
\usepackage[T1]{fontenc}
\usepackage[utf8]{inputenc}
\usepackage{microtype}
\usepackage{amssymb}
\usepackage{amsmath}
\usepackage[table]{xcolor}
\usepackage{graphicx}
\usepackage{subfig}
\usepackage{booktabs}
\usepackage{siunitx}
\usepackage{multirow}
\usepackage{enumitem}
\usepackage[bookmarks=false]{hyperref}
\hypersetup{colorlinks=true,linkcolor=blue,citecolor=blue,urlcolor=blue}

\setlist{nolistsep}
\title{\textbf{Counterfactual Reasoning for Robust Visual Question Answering}}
\author{Truong-Binh Duong$^{1,2}$, Thanh-Ngan Tran$^{1,2}$, Ngoc-Thao Nguyen$^{1,2}$, Bac Le$^{1,2,*}$\\[3pt]
\small $^1$Faculty of Information Technology, University of Science, Ho Chi Minh City, Vietnam\\
\small $^2$Vietnam National University, Ho Chi Minh City, Vietnam\\
\small $^*$Corresponding author: Bac Le (lhbac@fit.hcmus.edu.vn)}
\date{\small Accepted for publication at the 30th International Conference on Knowledge-Based and Intelligent Information \& Engineering Systems (KES 2026).}

\begin{document}
\maketitle
\vspace{-1.2em}

\begin{abstract}
Modern Visual Question Answering (VQA) models often exploit spurious correlations in training data, leading to poor out-of-distribution (OOD) generalization due to language bias. Although counterfactual learning has shown promise, existing methods can be improved to better guide attention toward causal evidence and strengthen feature discrimination. To address this, we propose a novel training framework that enhances counterfactual contrastive learning for VQA. Our framework introduces three key contributions: (1) a three-stage curriculum for stable multi-objective optimization, (2) an enhanced Batch-Contrastive loss for more discriminative feature learning, and (3) two novel regularizers, Answer-Contrastive (AC) loss to refine the prediction space and Gradient-Discrepancy (GD) loss to enforce causal visual grounding. Our model achieves a competitive accuracy of 61.64\% on the bias-sensitive VQA-CP v2 benchmark while maintaining 62.80\% on the standard VQA v2 dataset, yielding a small generalization gap of 1.16\%. This demonstrates a strong balance between OOD robustness and in-distribution performance.
\end{abstract}

\vspace{0.5em}

\section{Introduction}
\label{sec:introduction}
Visual Question Answering (VQA) is a key benchmark task in artificial intelligence, requiring models to comprehend both visual and textual information for complex reasoning \cite{antol2015vqa}. While modern architectures achieve strong performance on benchmarks such as VQA v2 \cite{goyal2017making}, this success often masks a critical limitation. Prior studies show these models exploit spurious correlations and language priors in training data, rather than performing genuine multimodal reasoning \cite{agrawal2016analyzing,cadene2019rubi,agrawal2018dont}. Consequently, performance degrades on out-of-distribution (OOD) datasets such as VQA-CP v2 \cite{agrawal2018dont}, designed to penalize such shortcuts. This raises a fundamental question: How can we train VQA models to reason from causal evidence in both the image and the question, instead of memorizing statistical artifacts?

Counterfactual learning has emerged as a promising direction. Methods such as Counterfactual Samples Synthesizing (CSS) \cite{chen2020counterfactual} generate augmented data by masking key elements in the image or question, thereby disrupting spurious correlations. Building on this idea, CL-VQA \cite{liang2020learning} pioneered the use of contrastive learning to explicitly model the relationships among original, factual, and counterfactual samples. This approach encourages the model to learn more generalizable representations by leveraging counterfactuals as a self-supervised signal. Despite these advances, existing methods still have limitations: the model’s attention is not explicitly guided toward causal regions, and the contrastive loss can be further improved to enhance discriminative power. 

To address the gaps, we propose an enhanced counterfactual contrastive learning framework. Our contributions are:
\begin{itemize}
    \item A novel three-stage curriculum training framework that systematically integrates multiple complex learning objectives, ensuring training stability and effective optimization.
    \item An enhanced contrastive learning paradigm featuring an improved batch-contrastive loss and two novel regularizers: an Answer-Contrastive (AC) Loss to refine the prediction space, and a Gradient-Discrepancy (GD) Loss to enforce causal visual grounding.
    \item Competitive performance on the VQA-CP v2 benchmark while maintaining a small generalization gap to the VQA v2 dataset.
\end{itemize}

\section{Related Work}

\subsection{Language Bias in VQA}
While modern VQA models demonstrate strong performance on in-distribution (ID) benchmarks like VQA v2 \cite{goyal2017making}, numerous studies \cite{agrawal2016analyzing,agrawal2018dont} have shown they exploit superficial linguistic correlations, or ``language biases'' in the training data. Instead of performing robust multimodal reasoning, models often learn to associate certain question patterns with high-frequency answers. For example, a model might learn to default to the answer ``tennis'' for any question beginning with ``What sport is...?'' regardless of the image content. This reliance on statistical shortcuts results in a substantial performance degradation on OOD test sets. To diagnose this vulnerability, the VQA-CP (Visual Question Answering under Changing Priors) dataset \cite{agrawal2018dont} was introduced, which intentionally creates a distribution shift between the training and test sets for each question type, making it a standard benchmark for evaluating model robustness.

\subsection{Debiasing Methods in VQA}
To address the language bias issue, one primary debiasing strategy uses ensemble-based models with an auxiliary, often question-only, branch to capture and regularize bias. These methods re-weight the loss or adjust predictions to downplay samples answerable from language priors, encouraging the main model to rely on visual evidence~\cite{cadene2019rubi,clark2019dont,han2021greedy,cho2023genb}. Other strategies modify the training objective, for instance by using adaptive margin losses to create more discriminative feature spaces~\cite{basu2023rmlvqa} or by improving visual grounding to ensure models are ``right for the right reasons''~\cite{wu2019self}. More recently, causal inference approaches have gained prominence in formally disentangling visual reasoning from linguistic shortcuts. Notable examples include CVIV+iter~\cite{pan2024unbiased}, which employs instrumental variables to isolate the causal effect of visual evidence on the answer, and CIBi~\cite{liu2024eliminating}, which applies fine-grained causal intervention to eliminate context and keyword biases separately.

Another major research line, most relevant to our work, focuses on data augmentation. It expands the training set with samples that break spurious correlations and create a more balanced distribution. A foundational technique is Counterfactual Samples Synthesizing (CSS) \cite{chen2020counterfactual}, which identifies causally critical elements, such as key objects in the image or essential words in the question, using gradient-based analysis. It then generates ``counterfactual'' data by masking these elements and assigning new, logically consistent answers. MUTANT \cite{gokhale2020mutant} extends this by generating ``mutant'' samples through semantic manipulations, such as changing an object's color via inpainting or negating a question's premise. By training the model to be consistent with these semantic shifts, it learns to understand the direct effect of input changes on the final answer. Unlike synthetic approaches, KDDAug \cite{chen2022rethinking} avoids potential generation artifacts by composing samples from existing pristine images and different human-written questions. This strategy assigns labels via knowledge distillation from pre-trained teachers, creating a scalable, rule-free approach.

However, simply generating more data is often insufficient because how the model utilizes these samples is crucial. This has led to contrastive learning frameworks, where CL-VQA~\cite{liang2020learning} first introduced a contrastive objective to learn the relationship between original (anchor), factual (positive), and counterfactual (negative) samples generated by CSS. Similarly, MMBS \cite{si2022towards} employs contrastive learning, constructing its positive samples by corrupting question-category information and treating biased and unbiased samples differently. Building on this foundation, we propose a curriculum and novel regularization losses to further strengthen the power of counterfactual contrastive learning for robust VQA.

\section{Proposed Method}
\label{sec:method}
We propose a training framework to enhance causal reasoning in VQA, built upon the UpDn backbone \cite{anderson2018bottom}. It integrates a counterfactual synthesis module and three novel loss functions within a three-stage curriculum.

\subsection{Baseline Architecture and Counterfactual Samples Synthesizing}

\begin{figure}[t]
    \centering
    \includegraphics[width=0.95\textwidth]{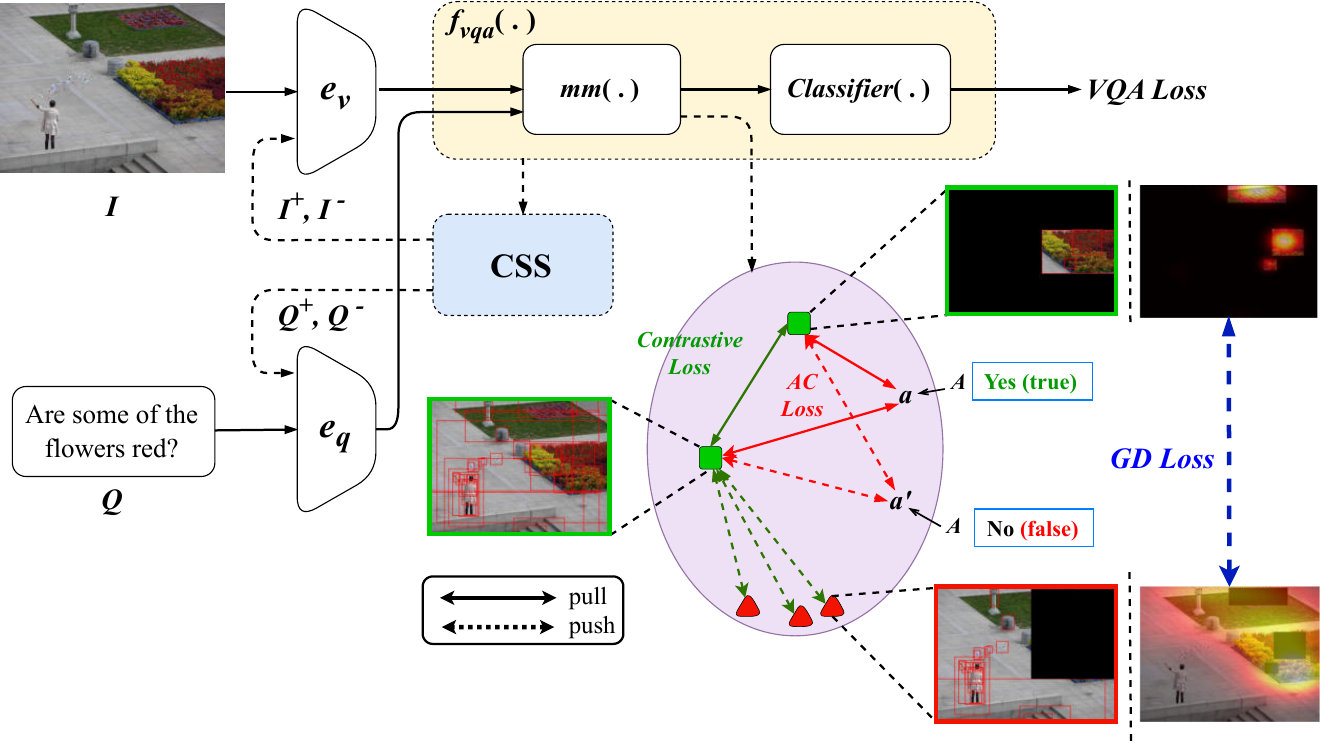}
    \caption{An overview of our training framework. The VQA baseline and CSS module generate feature triplets ($z, z^+, z^-$), optimized via a three-stage curriculum that progressively introduces our Batch-Contrastive ($\mathcal{L}_{\text{BC}}$), Answer-Contrastive ($\mathcal{L}_{\text{AC}}$), and Gradient-Discrepancy ($\mathcal{L}_{\text{GD}}$) losses.}
    \label{fig:method}
\end{figure}

\begin{figure}[t]
    \centering
    \includegraphics[width=0.8\linewidth]{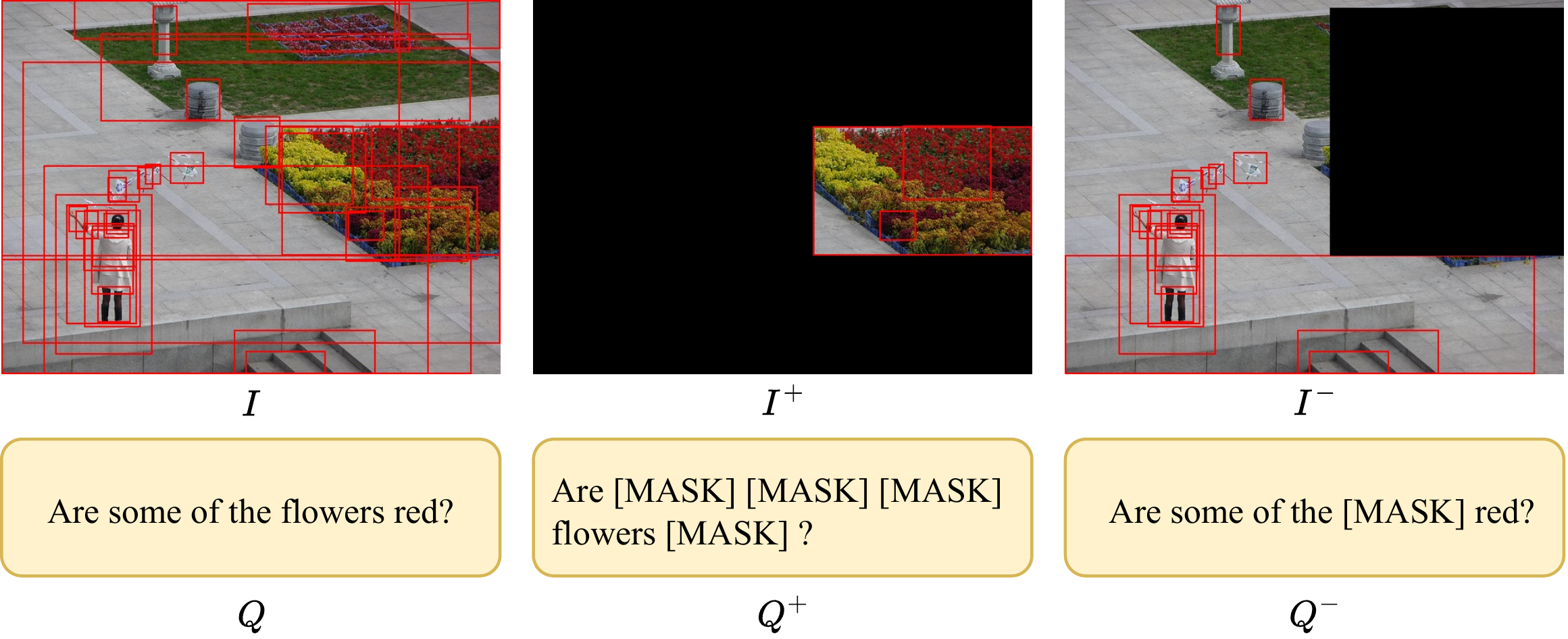}
    \caption{Example of factual ($I^+, Q^+$) and counterfactual ($I^-, Q^-$) samples generated by the CSS module, isolating and masking causal elements, respectively.}
    \label{fig:css}
\end{figure}

Formally, VQA learns a mapping $f_{vqa}: \mathcal{I} \times \mathcal{Q} \rightarrow \mathbb{R}^{|\mathcal{A}|}$ over dataset $\mathcal{D} = {(I_i, Q_i, a_i)}_{i=1}^N$, where $\mathcal{I}$, $\mathcal{Q}$, and $\mathcal{A}$ denote images, questions, and answers. Using the UpDn backbone \cite{anderson2018bottom}, as shown in Figure~\ref{fig:method}, the baseline extracts $k$ region features $V$ via image encoder $e_v$ and question features $q$ via text encoder $e_q$. A fusion module $mm(\cdot, \cdot)$ applies top-down attention to produce a joint feature $z = mm(V, q)$, mapped to answer logits $\hat{y} = C(z)$ by classifier $C$. Because the quality of $z$ dictates reasoning robustness, we employ contrastive learning. To mitigate spurious correlations, the CSS module \cite{chen2020counterfactual} acts as a data augmentation engine. For each sample $(I, Q, a)$, gradient-based analysis identifies critical regions or words. It synthesizes a factual sample ($I^{+}$ or $Q^{+}$) by retaining critical objects and a counterfactual sample ($I^{-}$ or $Q^{-}$) by masking them (Figure~\ref{fig:css}). Encoding these produces anchor ($z$), positive ($z^{+}$), and negative ($z^{-}$) features for contrastive optimization.

\subsection{Improved Loss Components}
\label{subsec:improved_losses}
Building on these causal triplets, we introduce three complementary loss functions targeting different reasoning aspects. They jointly structure the feature space, refine the prediction space, and ensure causal visual grounding.

\textbf{Batch-Contrastive Loss.}
The original CL-VQA objective \cite{liang2020learning} compares an anchor against single positive and negative samples. To learn a more discriminative representation, we adopt a Batch-Contrastive (BC) objective inspired by SimCLR \cite{chen2020simple}. This leverages all in-batch negatives for richer, more stable signals. Given a normalized anchor feature $z_i$, its positive $z_i^{+}$, and all negative features $\{z_k^{-}\}_{k=1}^N$ in a mini-batch of size $N$, the BC loss is:
\begin{equation}
\mathcal{L}_{\text{BC}}(i)
= - \log
\frac{e^{s(z_i, z_i^{+})/\tau}}{e^{s(z_i, z_i^{+})/\tau} + \sum_{k=1}^{N} e^{s(z_i, z_k^{-})/\tau}},
\label{eq:bcl}
\end{equation}
where $s(\cdot,\cdot)$ is the cosine similarity and $\tau > 0$ is a temperature hyperparameter scaling distribution sharpness. By contrasting one positive pair against numerous in-batch negatives, this objective supplies more robust gradients and forces the model to learn more generalizable features. 

\textbf{Answer-Contrastive Loss.}
While $\mathcal{L}_{\text{BC}}$ organizes the \textit{feature} space, it does not directly enforce separability in the final \textit{prediction} space. To address ambiguity between semantically similar answers (e.g., ``red'' vs. ``maroon''), we introduce an Answer-Contrastive (AC) Loss \cite{cho2023counterfactualmixup} operating on answer embeddings. Let $g(a)$ be the GloVe \cite{pennington2014glove} embedding of ground-truth answer $a \in \mathcal{A}$. For sample $i$ with ground-truth set $\mathcal{A}_i \subset \mathcal{A}$, the AC loss pulls both anchor ($z_i$) and positive ($z_i^{+}$) features closer to correct embeddings:
\begin{equation}
\mathcal{L}^{\mathrm{AC}}_{i}
= -
\log
\frac{
\sum_{a \in \mathcal{A}_i} \left( e^{s(z_i, g(a))} + e^{s(z_i^{+}, g(a))} \right)
}{
\sum_{a' \in \mathcal{A}} e^{s(z_i, g(a'))}
}.
\label{eq:ac_loss}
\end{equation}
The numerator aggregates similarity scores for all correct answers across both views, while the denominator normalizes over the entire vocabulary. This directly enlarges decision margins between correct and incorrect predictions.

\textbf{Gradient-Discrepancy Loss.}
Although BC and AC losses yield discriminative representations, they do not explicitly enforce correct visual grounding. The model might still exploit global statistical shortcuts rather than localizing causal evidence. To address this, we introduce a Gradient-Discrepancy (GD) Loss compelling the model to attend to relevant regions. It enforces a significant difference in model sensitivity, measured by output gradients with respect to visual features, between the factual ($z^{+}$) and counterfactual ($z^{-}$) views. Large discrepancy implies proper causal grounding, while low discrepancy implies shortcut learning, which we aim to penalize. To formalize this, we define the GD loss as a weighted objective that rewards gradient dissimilarity. Let $\hat{y}^{+}$ and $\hat{y}^{-}$ be the logits for positive and negative samples, $V_k$ the visual feature at region $k$, and $p_{\text{pos}}$ the positive pair probability from the BC head. The GD loss is formulated as:
\begin{equation}
\mathcal{L}_{\text{GD}}
=
-\mathbb{E}_{k}\!\left|
\frac{\partial \sum_{j}\hat{y}^{+}_{j}}{\partial V_{k}}
-
\frac{\partial \sum_{j}\hat{y}^{-}_{j}}{\partial V_{k}}
\right|
\cdot \log p_{\text{pos}}.
\label{eq:gd_loss}
\end{equation}

\begin{figure}[t]
    \centering
    \subfloat[High GD Loss (low discrepancy).]{\includegraphics[width=0.495\textwidth]{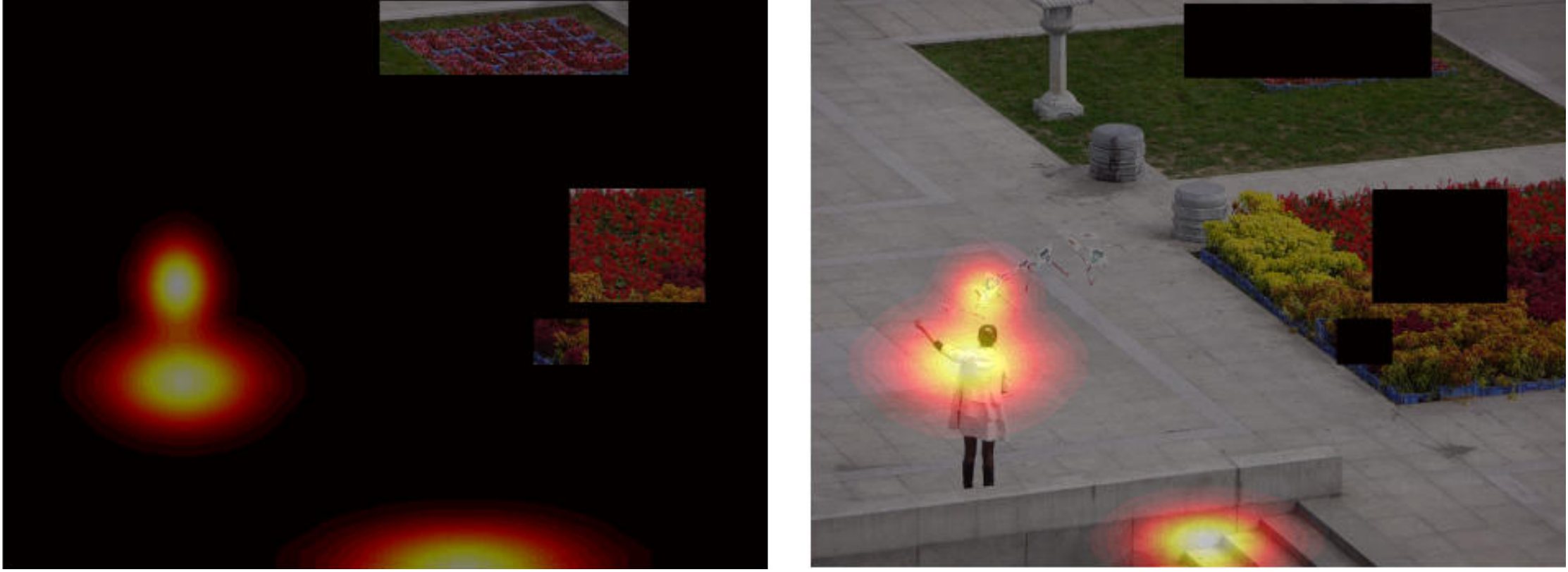}\label{fig:high_gd}}
    \hfill
    \subfloat[Low GD Loss (high discrepancy).]{\includegraphics[width=0.495\textwidth]{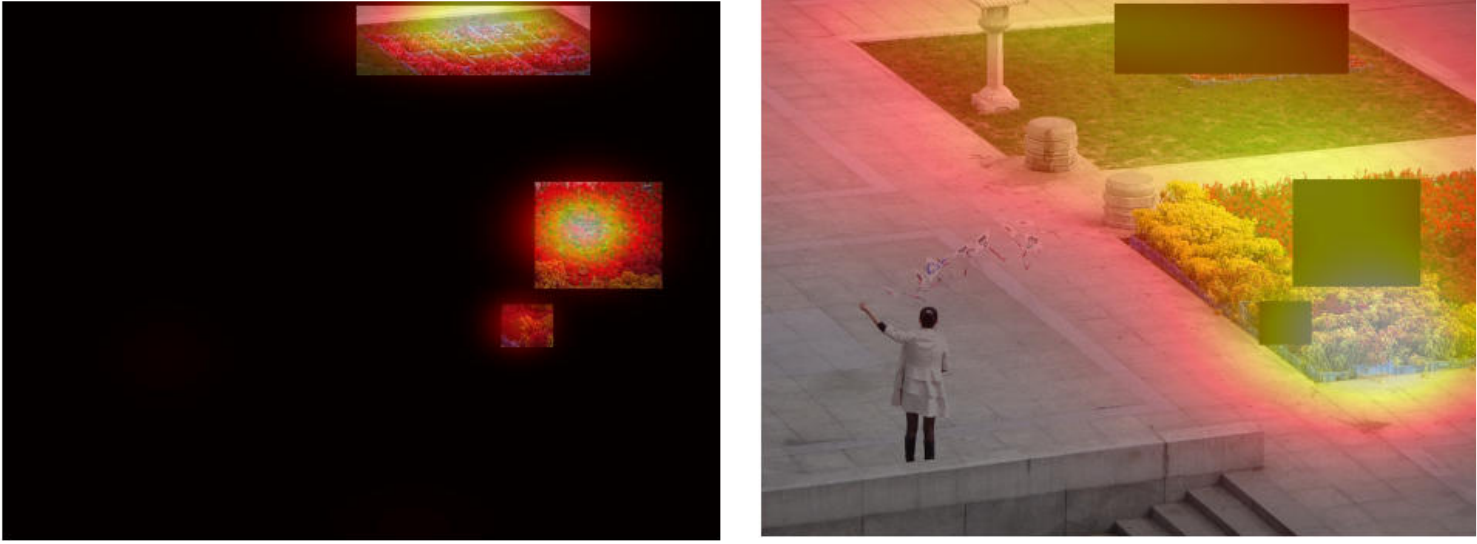}\label{fig:low_gd}}
    \caption{Visualization of the GD Loss. (a) A poorly-grounded model shows little attention discrepancy between factual and counterfactual sample, which in turn leads to a high loss. (b) A well-grounded model attends to causal evidence, producing high discrepancy and low loss.}
    \label{fig:gd_vis}
\end{figure}

This objective functions as a form of self-supervision for the model's attention. The absolute difference measures the region-wise gradient discrepancy, adaptively weighted by $\log p_{\text{pos}}$. Minimizing this loss penalizes low discrepancy, especially under uncertainty, forcing the model to develop distinct attention patterns. As illustrated in Figure~\ref{fig:gd_vis}, a well-grounded model (Fig.~\ref{fig:low_gd}) attends heavily to red flowers in $I^+$ and disperses attention when they are masked in $I^-$, yielding high discrepancy and low loss. Conversely, a biased model (Fig.~\ref{fig:high_gd}) maintains static attention on spurious cues, incurring a heavy penalty. This steers focus towards genuine causal evidence.

\subsection{Three-Stage Curriculum Training}
\label{subsec:curriculum}
Simultaneous optimization of complex objectives can destabilize training. We therefore adopt a three-stage curriculum strategy to progressively introduce loss components.

\noindent Stage 1 (Epochs 1--5) optimizes only the VQA task loss to establish stable base representations. Stage 2 (Epochs 6--15) adds BC to structure the embedding space around causal relationships. Stage 3 (Epochs 16--30) activates AC and GD to refine answer margins and enforce visual grounding:
\begin{equation}
\begin{aligned}
\mathcal{L}^{(1)} &= \mathcal{L}_{\text{VQA}},\\
\mathcal{L}^{(2)} &= \mathcal{L}_{\text{VQA}} + \lambda_{\text{BC}}\mathcal{L}_{\text{BC}},\\
\mathcal{L}^{(3)} &= \mathcal{L}_{\text{VQA}} + \lambda_{\text{BC}}\mathcal{L}_{\text{BC}} + \lambda_{\text{AC}}\mathcal{L}_{\text{AC}} + \lambda_{\text{GD}}\mathcal{L}_{\text{GD}}.
\end{aligned}
\end{equation}
The weighting coefficients $\lambda_{\text{BC}}$, $\lambda_{\text{AC}}$, and $\lambda_{\text{GD}}$ are reported in Section~\ref{sec:experiments} and analyzed in Table~\ref{tab:gd_sensitivity}. This schedule stabilizes optimization and incrementally strengthens causal reasoning.

\section{Experiments}
\label{sec:experiments}

\subsection{Experimental Settings and Implementation Details}
\label{subsec:experimental_settings}
We evaluate on VQA-CP v2 \cite{agrawal2018dont} for OOD robustness against language bias and VQA v2 \cite{goyal2017making} for ID performance. Following the standard fixed-split VQA-CP v2 protocol, we report the official VQA accuracy with \textit{Yes/No}, \textit{Number}, and \textit{Other} breakdowns. Cross-validation is not used because VQA-CP v2 is designed with different answer-prior distributions between training and test splits. Repartitioning the data would change the OOD setting and reduce comparability with prior work. We use the standard UpDn backbone \cite{anderson2018bottom} as a controlled testbed, train for 30 epochs with Adamax, batch size 512, gradient clipping 0.25, and cosine learning-rate decay after warm-up to $2 \times 10^{-3}$. The loss weights are constrained to a total budget of 10, with larger values indicating higher optimization priority. After repeated runs, we set $\lambda_{\text{BC}}=2.0$, $\lambda_{\text{AC}}=2.0$, and $\lambda_{\text{GD}}=6.0$, which produced the best overall trade-off in Table~\ref{tab:gd_sensitivity}. Code, model configurations, hyperparameter settings, and evaluation instructions are available at \href{https://github.com/duongtruongbinh/robust-vqa-counterfactual}{GitHub repository}.

\subsection{Quantitative Results}
\label{subsec:quantitative_results}
\begin{table}[!t]
\caption{Performance comparison with state-of-the-art methods on the VQA-CP v2 and VQA v2 datasets. Our Baseline refers to the CL-VQA configuration, while \textbf{Ours} incorporates the proposed BC, AC, and GD loss. The ``Gap'' column represents the absolute difference between the overall accuracies on VQA v2 and VQA-CP v2. $\checkmark$ denotes methods requiring extra annotations. $\uparrow$: higher is better, $\downarrow$: lower is better. Best results are in \textbf{bold}, second best are \underline{underlined}.}\label{tab:sota_comp}
\centering
\sisetup{detect-weight, detect-family, mode=text}
\setlength{\tabcolsep}{3pt}
\begin{tabular}{
    l 
    c 
    S[table-format=2.2]
    S[table-format=2.2]
    S[table-format=2.2]
    S[table-format=2.2]
    S[table-format=2.2]
    S[table-format=2.2]
    S[table-format=2.2]
    S[table-format=2.2]
    S[table-format=2.2]
}
\toprule
\textbf{Model} & \textbf{Extra} & \multicolumn{4}{c}{\textbf{VQA-CP v2 \textit{test} (\%) $\uparrow$}} & \multicolumn{4}{c}{\textbf{VQA v2 \textit{val} (\%) $\uparrow$}} & {\textbf{Gap $\downarrow$}} \\
\cmidrule(lr){3-6} \cmidrule(lr){7-10}
& \textbf{Anno.} & {\textbf{All}} & {\textbf{Y/N}} & {\textbf{Num}} & {\textbf{Other}} & {\textbf{All}} & {\textbf{Y/N}} & {\textbf{Num}} & {\textbf{Other}} & \\
\midrule
UpDn (2018) \cite{anderson2018bottom}& & 39.74 & 42.27 & 11.93 & 46.05 & \textbf{63.48} & \underline{81.18} & 42.14 & 55.66 & 23.74 \\
AdvReg (2018) \cite{ramakrishnan2018overcoming} & & 41.17 & 65.49 & 15.48 & 35.48 & 62.75 & 79.84 & 42.35 & 55.16 & 21.58 \\
RUBi (2019) \cite{cadene2019rubi} & & 44.23 & 67.05 & 17.48 & 39.61 & {-} & {-} & {-} & {-} & {-} \\
LMH (2019) \cite{clark2019dont}& & 52.01 & 72.58 & 31.12 & 46.97 & 56.35 & 65.06 & 37.63 & 54.69 & 4.34 \\
CSS (2020) \cite{chen2020counterfactual} & & 58.95 & 84.37 & 49.42 & 48.21 & 59.91 & 73.25 & 39.77 & 55.11 & 0.96 \\
CL-VQA (2020) \cite{liang2020learning}& & 59.18 & 86.99 & 49.89 & 47.16 & 57.29 & 67.27 & 38.40 & 54.71 & 1.89 \\
GGE (2021) \cite{han2021greedy} & & 57.32 & 87.04 & 27.75 & 49.59 & 59.11 & 73.27 & 39.99 & 54.39 & 1.79 \\
MMBS (2022) \cite{si2022towards} & & 56.44 & 76.00 & 43.77 & 49.67 & 61.87 & 75.86 & 40.34 & \textbf{56.95} & 5.43 \\
KDDAug (2022) \cite{chen2022rethinking} & & 61.14 & 88.31 & \textbf{56.10} & 48.28 & 62.17 & 79.50 & 40.57 & 54.71 & 1.03 \\
GenB (2023) \cite{cho2023genb} & & 59.15 & 88.03 & 40.05 & 49.25 & 62.74 & 86.18 & \textbf{43.85} & 47.03 & 3.59 \\
RMLVQA (2023) \cite{basu2023rmlvqa} & & 60.41 & \underline{89.98} & 45.96 & 48.74 & 59.99 & 76.68 & 37.54 & 53.26 & \textbf{0.42} \\
CIBi (2024) \cite{liu2024eliminating} & &  59.58 & 86.94 & 49.98 &  50.24 & 60.47 & 81.04 &  42.94 & 50.02 & 0.89 \\
CVIV+iter (2024) \cite{pan2024unbiased} & &  60.08 & 88.85 & 40.77 & \underline{50.30} & 61.42 & 79.34 & 39.85 & 53.50 & 1.34 \\
\textbf{Ours} & & \underline{61.64} & \textbf{90.44} & \underline{51.74} & 49.26 & 62.80 & 80.47 & 38.58 & \underline{55.77} & 1.16 \\
\midrule
HINT (2019) \cite{selvaraju2019taking}& $\checkmark$ & 46.73 & 72.36 & 10.61 & 45.88 & \underline{63.38} & 81.18 & \underline{42.99} & 55.56 & 16.65 \\
SCR (2019) \cite{wu2019self}& $\checkmark$ & 49.17 & 71.55 & 10.72 & 47.49 & 62.20 & 78.90 & 41.40 & 54.30 & 13.03 \\
MUTANT (2020) \cite{gokhale2020mutant}& $\checkmark$ & \textbf{61.72} & 88.90 & 49.68 & \textbf{50.78} & 62.56 & \textbf{82.07} & 42.52 & 53.28 & \underline{0.84} \\
\bottomrule
\end{tabular}
\end{table}

VQA debiasing aims to develop robustly generalizable models rather than specialists for a single OOD benchmark. An ideal method should achieve high accuracy on the bias-sensitive VQA-CP v2, maintain strong performance on the standard VQA v2, and exhibit a minimal generalization gap. Table~\ref{tab:sota_comp} evaluates our method against these criteria. Our model achieves \textbf{61.64\%} on VQA-CP v2 and \textbf{62.80\%} on VQA v2, with a small 1.16\% generalization gap. Among annotation-free methods, it obtains the highest overall VQA-CP v2 accuracy and the best Yes/No score, while remaining competitive on VQA v2. Compared with MUTANT~\cite{gokhale2020mutant}, which uses extra annotations, our method reaches a similar OOD accuracy without additional annotation resources. These results support the effectiveness of the proposed objectives, while the remaining gap to some recent methods suggests room for further improvement.

\subsection{Qualitative Results}
\label{subsec:qualitative_results}

\begin{figure}[!t]
    \centering
    \includegraphics[width=0.75\linewidth]{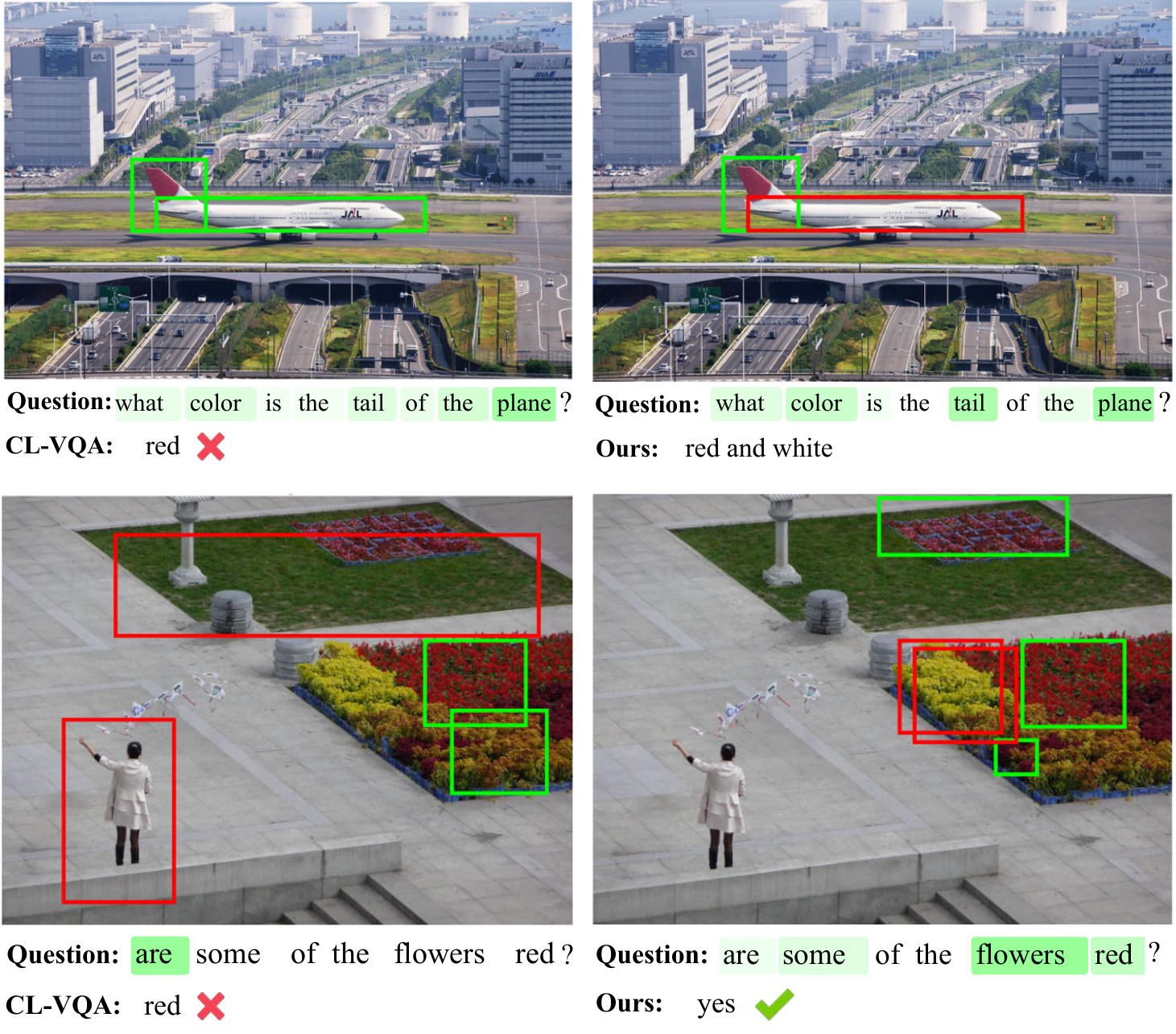}
    \caption{Qualitative comparisons between our model and CL-VQA. Green boxes mark regions that support the prediction, red boxes denote regions being suppressed, and word-level importance is visualized with different shades of green.}
    \label{fig:qualitative}
\end{figure}

To illustrate how our method improves reasoning, we qualitatively compare our model against the CL-VQA baseline (Figure~\ref{fig:qualitative}). Following CSS~\cite{chen2020counterfactual}, we visualize the contribution of image regions and question words to the predicted answer. Green boxes denote positive contributions, red boxes denote suppressive ones, and darker green text indicates higher word importance. The top example highlights improved grounding. The baseline provides an incomplete answer (``red'') with diffused attention across the airplane. In contrast, our model correctly predicts ``red and white'' by precisely concentrating attention on the tail and suppressing other parts. Linguistically, focusing on ``tail'' and ``plane'' reflects superior semantic grounding. The bottom example demonstrates robustness against language priors. Faced with a yes/no question, the baseline succumbs to bias by answering with a color (``red'') and misdirecting attention. Conversely, our model correctly answers ``yes'' by grounding reasoning in actual visual evidence. It accurately encompasses the red flowers while suppressing distractors like yellow flowers, with linguistic attention correctly centered on the causal terms ``flowers'' and ``red''. Collectively, these examples demonstrate that the AC and GD losses successfully guide the model toward causal evidence for accurate and well-grounded reasoning.

\subsection{Analysis of Answer Distribution}
\label{subsec:distribution_analysis}

\begin{figure}[!t]
    \centering
    \includegraphics[width=0.95\textwidth]{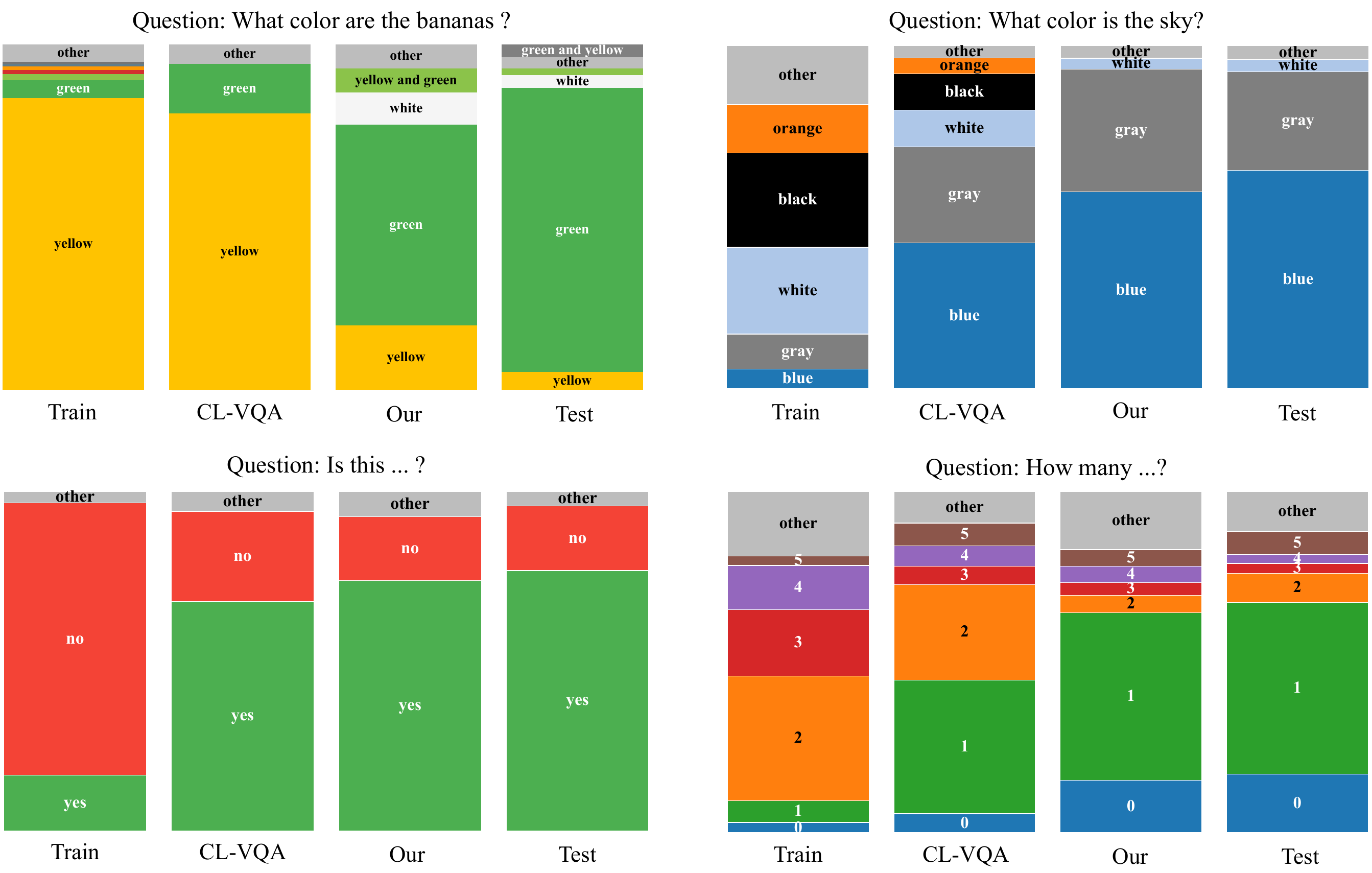}
    \caption{Comparison of answer distributions predicted by our model and the CL-VQA baseline, alongside the ground-truth distributions of the train and test sets of VQA-CP v2 for several representative question types.}
    \label{fig:distribution_analysis}
\end{figure}

To provide direct evidence of bias mitigation, we analyze predicted answer distributions on the VQA-CP v2 test set (Figure~\ref{fig:distribution_analysis}). The analysis confirms our model adapts to the test distribution rather than memorizing training-set biases. For the question \textit{``What color are the bananas?''}, the training data is skewed toward ``yellow''. Unlike the baseline, our model correctly aligns with the test set's ``green'' majority. This trend holds across other categories. For \textit{``Is this...?''} questions, our model successfully overcomes the training bias toward ``no'' to reflect the test set's ``yes'' majority. These findings confirm that our framework reduces reliance on spurious language correlations, enabling better generalization under distribution shifts.

\subsection{Ablation Studies}
\label{subsec:ablation}
To isolate the contribution of each component, we conduct ablation studies on VQA-CP v2 evaluating our loss functions, their weight sensitivity, and the curriculum strategy.

\textbf{Analysis of Loss Components.}
Table~\ref{tab:ablation_ac_gd} details the impact of the AC and GD losses built upon two distinct baselines (UpDn+LMH+CSS and our UpDn+LMH+CSS+BC). On the CSS baseline, both AC and GD provide modest but inconsistent improvements. GD reduces performance on Number questions, and their combination yields no further gains over using GD alone. This suggests these regularizers require the stabilizing effect of a stronger contrastive objective to integrate synergistically. This dynamic changes significantly when combined with our BC loss. A clear synergistic effect emerges. GD delivers a substantial 4-point boost to Y/N accuracy (90.52\%), while AC proves crucial for Number questions (53.24\%). Their combination achieves the highest overall accuracy of \textbf{61.64\%}, confirming that AC and GD play complementary roles unlocked by the discriminative feature space of our BC loss.

\setlength{\tabcolsep}{10pt}
\begin{table}[!t]
\caption{Individual and synergistic impact of the AC and GD loss on the VQA-CP v2 test dataset. The BC (Batch-Contrastive) loss refers to our improved contrastive formulation.}
\label{tab:ablation_ac_gd}
\centering
\begin{tabular}{lcccc}
\toprule
\textbf{Methods} & \textbf{All} & \textbf{Y/N} & \textbf{Number} & \textbf{Other} \\ 
\midrule
UpDn+LMH & 52.01 & 72.58 & 31.12 & 46.97 \\
\midrule
+ CSS & 58.95 & 84.37 & 49.42 & 48.21 \\
\midrule
\quad + AC & 59.54 & 84.06 & 50.91 & 49.05 \\
\quad + GD & 59.86 & 87.49 & 45.17 & \textbf{49.40} \\
\quad + AC + GD & 59.67 & 86.62 & 45.71 & 49.37 \\
\midrule
+ CSS+BC & 59.98 & 86.11 & 51.36 & 48.66 \\
\midrule
\quad + AC & 60.57 & 87.09 & \textbf{53.24} & 48.52 \\
\quad + GD & 61.41 & \textbf{90.52} & 51.47 & 48.89 \\
\quad + AC + GD & \textbf{61.64} & 90.44 & 51.74 & 49.26 \\
\bottomrule
\end{tabular}
\end{table}

\textbf{Sensitivity to GD Loss Weight.}
Table~\ref{tab:gd_sensitivity} analyzes the sensitivity of the GD loss weight $\lambda_{\text{GD}}$. Setting 2 increases the emphasis on causal grounding by raising $\lambda_{\text{GD}}$ to 6.0 while decreasing $\lambda_{\text{AC}}$ to 2.0, achieving a better overall trade-off. While Setting 1 peaks on the Y/N metric, Setting 2 improves the Number, Other, and overall OOD accuracy to \textbf{61.64\%}. This stronger grounding signal also yields superior ID performance, surpassing both Setting 1 and the MUTANT baseline on VQA v2 overall and Other accuracies.

\setlength{\tabcolsep}{4pt}
\begin{table}[!t]
\caption{Analysis of GD loss weight ($\lambda_{\text{GD}}$). Ours (Setting 2) showcases the best overall trade-off between OOD robustness and ID generalization.}
\label{tab:gd_sensitivity}
\centering
\begin{tabular}{l ccc | cccc | cccc}
\toprule
\multicolumn{1}{c}{\multirow{2}{*}{\textbf{Model}}} & \multicolumn{3}{c}{\textbf{Weights}} & \multicolumn{4}{c}{\textbf{VQA-CP v2 test (\%)}} & \multicolumn{4}{c}{\textbf{VQA v2 val (\%)}} \\
\cmidrule(lr){2-4} \cmidrule(lr){5-8} \cmidrule(lr){9-12}
& $\lambda_{\text{BC}}$ & $\lambda_{\text{AC}}$ & $\lambda_{\text{GD}}$ & \textbf{All} & \textbf{Y/N} & \textbf{Num} & \textbf{Other} & \textbf{All} & \textbf{Y/N} & \textbf{Num} & \textbf{Other} \\
\midrule
MUTANT \cite{gokhale2020mutant} & - & - & - & \textbf{61.72} & 88.90 & 49.68 & \textbf{50.78} & 62.56 & \textbf{82.07} & \textbf{42.52} & 53.28 \\
\midrule
Ours (Setting 1) & 2.0 & 4.0 & 4.0 & 61.41 & \textbf{90.52} & 51.47 & 48.89 & 62.43 & 79.62 & 38.46 & 55.75 \\
Ours (Setting 2) & 2.0 & 2.0 & 6.0 & 61.64 & 90.44 & \textbf{51.74} & 49.26 & \textbf{62.80} & 80.47 & 38.58 & \textbf{55.77} \\
\bottomrule
\end{tabular}
\end{table}

\textbf{Effect of the Improved Contrastive Loss.}
Table~\ref{tab:ablation_contras_loss} validates the importance of our BC loss against the original contrastive loss (CL) from CL-VQA~\cite{liang2020learning}. Substituting CL with BC increases overall accuracy from 60.82\% to \textbf{61.64\%}. This underscores that leveraging harder negative sampling via a batch-wise objective is crucial for learning discriminative representations.

\setlength{\tabcolsep}{8pt}
\begin{table}[!t]
\caption{Comparison of the original contrastive loss (CL) against our BC loss. Both are based on the full framework (UpDn+LMH+CSS+AC+GD).}\label{tab:ablation_contras_loss}
\centering
\sisetup{detect-weight, detect-family, mode=text}
\begin{tabular}{lS[table-format=2.2]S[table-format=2.2]S[table-format=2.2]S[table-format=2.2]}
\toprule
\multicolumn{1}{c}{\textbf{Method}} & {\textbf{All}} & {\textbf{Y/N}} & {\textbf{Number}} & {\textbf{Other}} \\
\midrule
Full Framework with CL & 60.82 & 89.51 & 48.59 & 49.13 \\
Full Framework with BC (Ours) & \textbf{61.64} & \textbf{90.44} & \textbf{51.74} & \textbf{49.26} \\
\bottomrule
\end{tabular}
\end{table}

\textbf{Effect of Curriculum Training.}
Table~\ref{tab:ablation_curriculum} evaluates our staged training strategy. Training all components jointly from the start yields a poor 52.94\% accuracy due to unstable multi-objective optimization. A two-stage schedule mitigates this issue, improving the score to 61.39\%. However, our proposed three-stage curriculum yields the best result of \textbf{61.64\%}, confirming that a gradual introduction of objectives ensures stable training and robust generalization.

\setlength{\tabcolsep}{6pt}
\begin{table}[!t]
\caption{Impact of the Three-Stage Curriculum Training strategy.}
\label{tab:ablation_curriculum}
\centering
\begin{tabular}{p{4.6cm} p{2cm} p{2.5cm}|cccc}
\toprule
\multicolumn{3}{c}{\textbf{Training Stages}} & \multicolumn{4}{c}{\textbf{VQA-CP v2 \textit{test} (\%)}} \\
\cmidrule(lr){1-3} \cmidrule(lr){4-7}
\cellcolor{blue!20}\centering\shortstack{\textbf{Stage 1}\\(Epochs 1--5)} &
\cellcolor{green!20}\centering\shortstack{\textbf{Stage 2}\\(Epochs 6--15)} &
\cellcolor{orange!20}\centering\shortstack{\textbf{Stage 3}\\(Epochs 16--30)} &
\textbf{All} & \textbf{Y/N} & \textbf{Number} & \textbf{Other} \\
\midrule
\cellcolor{blue!20}VQA Loss + BC + AC + GD & \cellcolor{blue!20} & \cellcolor{blue!20} & 52.94 & 71.93 & 42.18 & 45.94 \\
\cellcolor{blue!20}VQA Loss + BC & \cellcolor{green!20}+AC+GD & \cellcolor{green!20} & 61.39 & 90.41 & 51.49 & 48.91 \\
\cellcolor{blue!20}VQA Loss & \cellcolor{green!20}+BC & \cellcolor{orange!20}+AC+GD & \textbf{61.64} & \textbf{90.44} & \textbf{51.74} & \textbf{49.26} \\
\bottomrule
\end{tabular}
\end{table}

\subsection{Threats to Validity}
\label{subsec:threats}
Our evaluation follows the fixed VQA-CP v2/VQA v2 splits for comparability, but results may vary with initialization and implementation details. Moreover, VQA-CP v2 mainly tests answer-prior shifts and cannot represent all OOD conditions. We instantiate the framework on UpDn to isolate the contribution of the proposed objectives. Whether the same gains transfer to newer architectures should be further examined. Reproducibility also depends on dataset preprocessing, external image features, and software versions, which are documented with the released code.

\section{Conclusion}
\label{sec:conclusion}
We introduce a training framework to enhance causal reasoning in VQA via counterfactual contrastive learning. Our approach combines an improved Batch-Contrastive loss with Answer-Contrastive and Gradient-Discrepancy losses within a stable three-stage curriculum. The model achieves 61.64\% accuracy on VQA-CP v2 and 62.80\% on VQA v2, with a small generalization gap. Ablation studies show this performance stems from the complementary roles of each component in refining the feature space, prediction space, and visual grounding. This work presents a robust approach towards more reliable VQA models that balance OOD robustness and ID performance. Future work includes applying this framework to larger Transformer-based architectures to further advance multimodal causal reasoning.

\section*{Acknowledgements}
This research is funded by Vietnam National University, HoChiMinh City (VNU-HCM) under grant number CB2025-18-04.

\end{document}